\documentclass[10pt,twocolumn,letterpaper]{article}

\usepackage[pagenumbers]{wacv} 
\usepackage{xcolor}
\usepackage{graphicx}
\usepackage{multirow}
\usepackage[pagebackref]{hyperref}
\usepackage{amsmath}
\usepackage{xspace}
\usepackage{subcaption}
\usepackage{rotating}
\usepackage[table]{xcolor}
\usepackage{colortbl}
\usepackage{bbm}
\newcommand{\name}[0]{{\sc BoundarySeg }\xspace}
\usepackage{amsmath}
\usepackage{placeins}
\usepackage[utf8]{inputenc}

\definecolor{wacvblue}{rgb}{0.21,0.49,0.74}
\usepackage{hyperref}

\def\wacvPaperID{****} 
\def\confName{WACV}
\def\confYear{2027}

\title{\name: Boundary-Aware Multi-Task Learning for Semi-Supervised 3D Medical Image Segmentation with Limited Unlabeled Data}

\author{
Tushar Kataria \qquad Shireen Y. Elhabian\\
{\tt\small \{tushar.kataria,shireen\}@sci.utah.edu}\\
Scientific Computing and Imaging Institute \& Kahlert School of Computing\\
University of Utah, Salt Lake City, UT, USA
}

\begin{document}
\maketitle
\begin{abstract}
Data scarcity is a primary bottleneck in 3D medical image segmentation: expert voxel-level annotations are expensive, and privacy regulations often severely restrict access to unannotated clinical volumes. While semi-supervised learning (SSL) aims to alleviate annotation burdens, existing paradigms implicitly rely on large pools of unlabeled data ($U$) relative to labeled data ($L$), assuming $U \gg L$ to filter noise and enforce consistency. When restricted to the data-constrained regime where unlabeled data are limited ($U \le L$), standard SSL methods suffer from severe pseudo-label degeneracy, performance degradation, and high seed-to-seed variance.
In this work, we propose \name, a unified multi-task semi-supervised framework specifically designed for the limited unlabeled data regime ($U \le L$). \name formulates volumetric segmentation as a joint learning task that predicts whole-organ geometry alongside auxiliary organ boundaries. Rather than relying on large cohorts to average out pseudo-label noise, \name employs the auxiliary boundary branch as an internal geometric anchor. Coupled with uncertainty-weighted mean-teacher consistency across both tasks alongside axis-flip equivariance and copy-paste regularization, this joint supervision stabilizes spatial representations and prevents segmentation collapse when $U \le L$. Across multiple 3D segmentation datasets, \name maintains high performance stability as unlabeled pool size shrinks, consistently outperforming state-of-the-art semi-supervised frameworks in low-data settings. \href{https://github.com/tushaarkataria/BoundarySeg}{https://github.com/tushaarkataria/BoundarySeg}
\end{abstract}
    
\section{Introduction}
\label{sec:intro}

Deep learning has achieved remarkable success across 3D medical image segmentation tasks, including organ parsing \cite{wang2023one,rehrl2021evaluating,ronneberger2015u}, tumor localization \cite{dexl2026autopet3challengeautomatedlesion,shaker2024unetr++,li2025pantspancreatictumorsegmentation}, and automated clinical decision support \cite{dahiwade2019designing,ozturk2020automated}. However, translating these models into clinical workflows requires massive, high-quality training datasets \cite{litjens2017survey,li2024abdomenatlas}. In medical imaging, crowd-sourcing annotations is impossible; generating pixel- or voxel-level masks requires domain expertise from radiologists and clinicians \cite{willemink2020preparing,budd2021survey,li2024abdomenatlas}. This bottleneck is exacerbated in 3D volumetric scans, where annotating dense anatomical boundaries across dozens of slices is labor-intensive, costly, and difficult to scale \cite{tajbakhsh2020embracing,zheng2025significant,li2024abdomenatlas,yeung2021sli2vol}.

\begin{figure*}[!htb]
    \centering
    \includegraphics[width=0.95\linewidth]{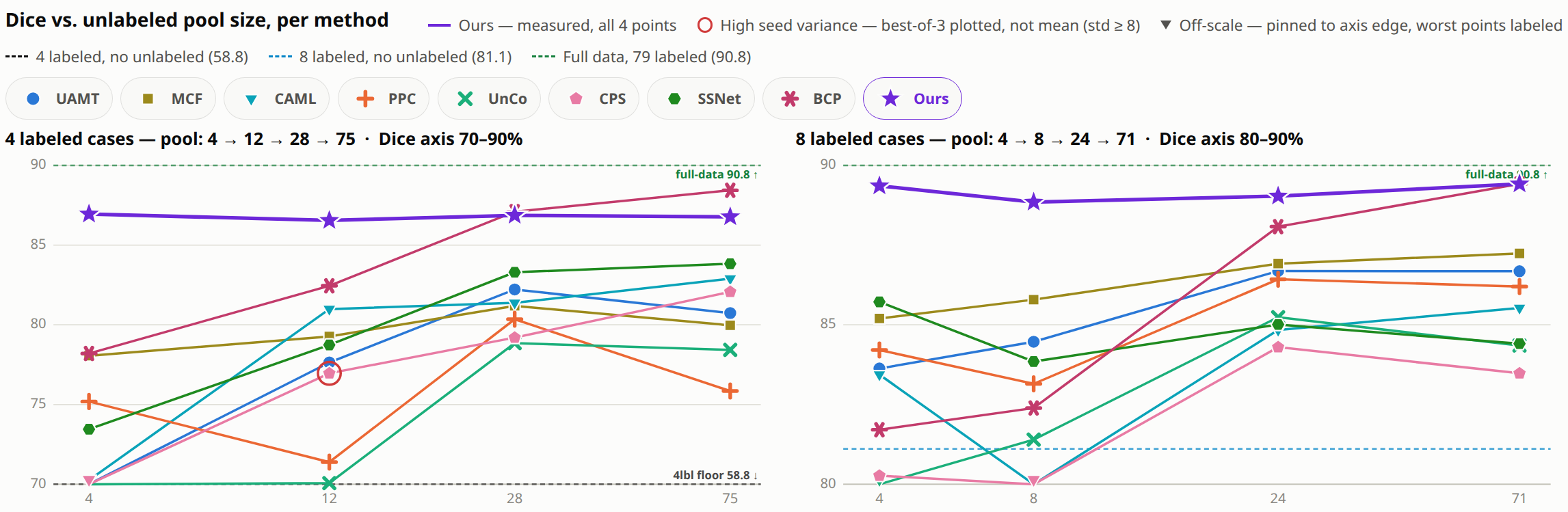}
    \caption{\textbf{Semi-supervised methods depend on unlabeled data volume for left atrium segmentation.} We fix the number of labeled MRI volumes at 4 (left) or 8 (right) and vary the unlabeled pool from 4 to 75 cases (with the same 79-case pool split differently in each panel). We compare eight semi-supervised baselines (UAMT, MCF, CAML, PPC, UnCo, CPS, SSNet, BCP) and our method over 3 seeds; markers show mean Dice. Dashed lines indicate the supervised floor and the full-data ceiling. Most baselines drop by 5--20 Dice points as unlabeled data shrink, while our method stays nearly constant across all pool sizes.}
    \label{fig:motivation_figure}
    \vspace{-1.5em}
\end{figure*}

 A second challenge is that medical data, whether annotated or unannotated, are difficult to obtain at scale because of privacy regulations and institutional data-protection constraints \cite{kaissis2020secure,rieke2020future,teo2024federated,pati2024privacy}. Regulations such as the Health Insurance Portability and Accountability Act (HIPAA) govern how protected health information may be accessed, shared, and reused, often requiring patient authorization, institutional approval, data-use agreements, or appropriate de-identification \cite{hhs_hipaa_deidentification}. Consequently, access to both labeled and unlabeled data remains a practical constraint in many medical-imaging studies, particularly those involving specialized anatomical structures, private clinical cohorts, or institution-specific acquisition protocols \cite{kaissis2020secure,willemink2020preparing}. This constraint exposes a key limitation of current semi-supervised learning methods: their reliance on a sufficiently large pool of unlabeled data. Semi-supervised segmentation methods aim to reduce annotation requirements by combining a small labeled set with a substantially larger collection of unlabeled volumes. Using strategies such as pseudo-labeling \cite{wang2023mcf}, consistency regularization \cite{wu2022mutual,wu2022exploring}, strong augmentation \cite{bai2023bidirectional,yan2025sgtc}, auxiliary geometric supervision \cite{li2020shape,luo2021semi,hu2025mtcl}, and uncertainty-aware mean-teacher frameworks \cite{yu2019uncertainty,zeng2025uncertainty}, these methods have achieved strong performance when abundant unlabeled data are available. Their improvements, however, depend heavily on the size and diversity of this unlabeled pool. When unlabeled volumes are scarce, the advantage over purely supervised training can diminish substantially, as illustrated in Figure~\ref{fig:motivation_figure}. Although semi-supervised learning reduces the need for manual annotation, it does not fully resolve the challenge of data scarcity in privacy-constrained medical imaging.

To resolve this failure mode, we propose \name, a multi-task semi-supervised framework specifically designed to operate robustly in the data-constrained regime ($U \le L$). Rather than relying on large unlabeled cohorts to filter pseudo-label noise, \name extracts maximum spatial regularization directly from available training data by jointly predicting whole-organ masks and auxiliary organ boundaries automatically derived from ground-truth labels. In low-unlabeled regimes ($U \le L$), pseudo-labels easily suffer from feature blur and noise amplification \cite{tarvainen2017mean,arazo2020pseudo}, so the auxiliary boundary prediction acts as an internal geometric anchor\cite{luo2021semi,li2020shape}, penalizing structural deviations and preserving feature representations across unannotated scans. We integrate this multi-task structure into an uncertainty-weighted mean-teacher framework \cite{tarvainen2017mean,yu2019uncertainty} and combine it with axis-flip equivariance and copy paste augmentation as regularization \cite{cirillo2021best} to enforce spatial consistency across symmetric transformations. While components such as auxiliary boundary supervision and mean-teacher consistency exist independently in the literature, current SSL frameworks deploy them under the assumption of large unlabeled pools ($U \gg L$). The novelty of \name lies in the unified design of a large unlabeled data-independent multi-task pipeline: by coupling uncertainty-weighted dual-head consistency with transformation equivariance , \name converts boundary learning into a stabilizing geometric anchor that prevents pseudo-label collapse specifically when $U \le L$. This provides robust supervision without requiring computationally expensive modules or extra overhead. Our main contributions are summarized as follows:
\begin{itemize}
    \item \textbf{Unlabeled Data Constrained Regime ($U \le L$):} We train and evaluate the limited unlabeled data regime ($U \le L$)  revealing severe performance degradation for 3D medical image semi-supervised segmentation (\Cref{fig:motivation_figure}).
    
    \item \textbf{\name Framework:} We propose a unified multi-task framework that integrates auxiliary boundary supervision as an internal geometric anchor, coupled with uncertainty-weighted dual-head consistency and axis-flip equivariance \& copy paste augmentation as regularization to stabilize representations under data scarcity.
    
    \item We show that \name achieves better performance across multiple datasets for $U \le L$. 
\end{itemize}
\section{Related Works}
\label{sec:related_works}

\begin{table*}[!htb]
\centering
\caption{\textbf{Conceptual positioning of \name.} Standard SSL relies on large unlabeled cohorts ($U \gg L$) for noise filtering and consistency. Conversely, \name leverages ground-truth supervision to stabilize representations when unlabeled data is scarce.}
\label{tab:method_positioning}
\resizebox{\linewidth}{!}{
\begin{tabular}{lllc}
\toprule
Category & Representative Methods & Core Mechanism & Robust when $U \le L$? \\
\midrule
Uncertainty \& Consistency
& UA-MT \cite{yu2019uncertainty}, UnCo \cite{zeng2025uncertainty}
& Teacher-student prediction variance and confidence thresholding
& No \\

Mutual / Decoder Disagreement
& MC-Net \cite{wu2022mutual}, SS-Net \cite{wu2022exploring}
& Cross-decoder feature disagreement and spatial smoothness
& No \\

Pseudo-Label Refinement
& MCF \cite{wang2023mcf}
& Subnetwork mutual correction and edge-guided pseudo-labeling
& No \\

Geometry-Aware SSL
& SASSNet \cite{li2020shape}, DTC \cite{luo2021semi}, MTCL \cite{hu2025mtcl}
& Distance field (SDM) regression and level-set consistency
& No \\

Augmentation \& Co-Training
& BCP \cite{bai2023bidirectional}, SGTC \cite{yan2025sgtc}
& Bidirectional copy-paste and sparse-slice co-training
& No \\

\midrule
\textbf{Boundary-Anchored Multi-Task}
& \textbf{\name (Ours)}
& \textbf{Boundary classification anchor, dual consistency, flip equivariance}
& \textbf{Yes} \\
\bottomrule
\end{tabular}}
\vspace{-1em}
\end{table*}

\textbf{Semi-Supervised 3D Medical Segmentation.} Semi-supervised learning (SSL) aims to leverage unannotated volumes to improve segmentation accuracy under limited manual annotations \cite{litjens2017survey,willemink2020preparing}. Prior approaches predominantly rely on consistency regularization, teacher-student architectures, or pseudo-label refinement. Uncertainty-aware mean-teacher (UA-MT) \cite{yu2019uncertainty} and UnCo \cite{zeng2025uncertainty} filter unconfident predictions using Monte Carlo rollouts or variance thresholds. MC-Net \cite{wu2022mutual} and SS-Net \cite{wu2022exploring} utilize multi-decoder networks to enforce cross-decoder disagreement and feature smoothness. MCF \cite{wang2023mcf} refines pseudo-labels by correcting model cognitive bias between paired networks. Other methods incorporate strong spatial augmentations or co-training strategies; BCP \cite{bai2023bidirectional} uses bidirectional copy-paste compositing across labeled and unlabeled volumes, while SGTC \cite{yan2025sgtc} combines sparse-slice annotations with semantic-guided co-training. 

While effective when unannotated data are abundant ($U \gg L$), these paradigms share a fundamental assumption: that early prediction errors and noise can be averaged out over a large, diverse unlabeled pool. When restricted to the data-constrained regime ($U \le L$), this assumption breaks down. Without sufficient cohort diversity, noisy teacher predictions reinforce confirmation bias \cite{arazo2020pseudo,tarvainen2017mean}, leading to pseudo-label degeneracy and high seed-to-seed variance. As summarized in Table~\ref{tab:method_positioning}, \name addresses this vulnerability by building supervision that remains effective even when $U \le L$.

\textbf{Multi-Task and Geometry-Aware Learning.} Multi-task learning (MTL) improves representation learning by sharing lower-level features across complementary objectives, serving as an implicit regularizer against overfitting \cite{zhang2021survey,sener2018multi,fifty2021efficiently,kataria2024staindiffuser,standley2020tasks}. In 3D medical image segmentation, auxiliary tasks providing geometric priors—such as distance transform maps, level sets, or boundary maps—are widely deployed to enforce spatial plausibility and regularize regional predictions \cite{murugesan2019psi,kervadec2021boundary,jurdi2021surprisingly,stewart2023regression}. Forcing the backbone to encode both regional semantics and surface contours guides the network toward feature representations that resist noise along low-contrast boundaries. Geometry-aware SSL methods such as SASSNet \cite{li2020shape}, DTC \cite{luo2021semi}, and MTCL \cite{hu2025mtcl} build on this principle by regressing continuous Signed Distance Maps (SDMs) alongside region segmentation to constrain unlabeled predictions.

However, pairing regional classification objectives (e.g., Dice and cross-entropy) with spatial distance regression (e.g., $L_1$/$L_2$ SDM loss) introduces optimization challenges due to mismatched loss scales and gradient dynamics \cite{yu2020gradient,chen2018gradnorm,wang2020gradient,stewart2023regression}. Continuous distance fields also rely on sensitive truncation thresholds and scale-specific normalization \cite{phan2025beyond}, which can exacerbate optimization instability when unlabeled data are scarce. To eliminate these cross-task mismatches, \name reformulates geometric supervision strictly within the \emph{classification space}. By extracting explicit binary organ boundaries via 3D mathematical morphology (erosion and XOR), both the primary organ head and auxiliary boundary head share identical, scale-invariant loss functions (Dice + Cross-Entropy) and uncertainty-weighted consistency schedules. This unified classification design avoids loss re-balancing and gradient conflicts, allowing the boundary task to function as a robust, hyperparameter-light geometric anchor that stabilizes spatial representations when $U \le L$.
\section{Methods : \name}\label{sec:formatting}

\begin{figure*}[!htb]
    \centering
   \includegraphics[  trim={0.5cm 7.0cm 1.9cm 2.4cm},
        clip=true,
        width=0.9\linewidth]{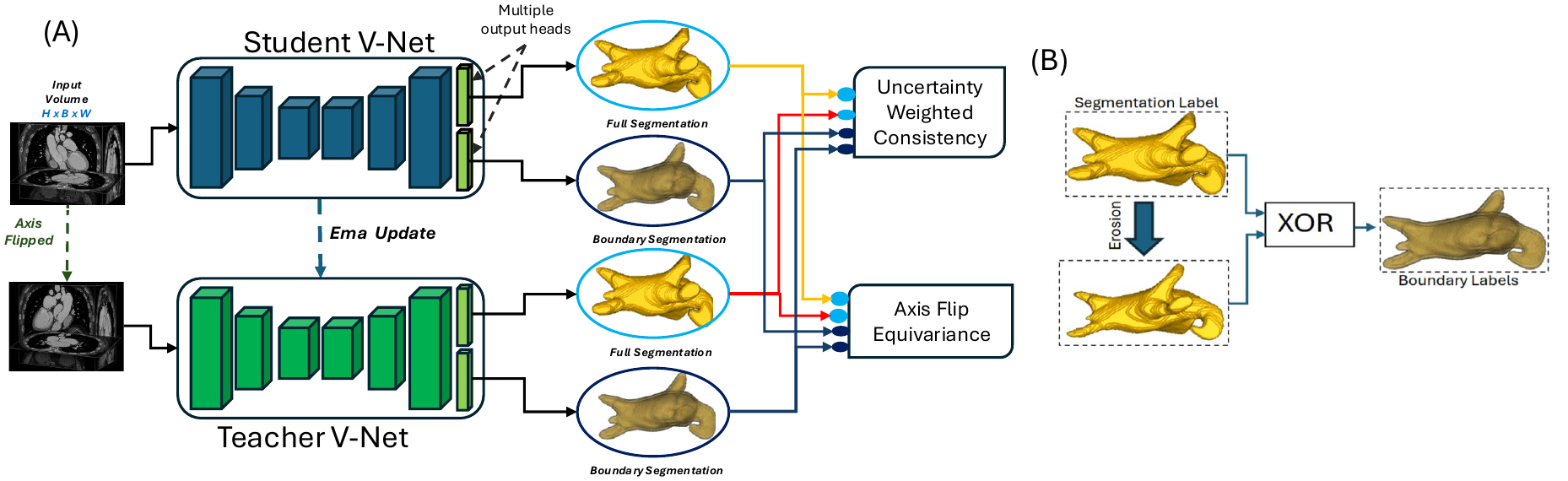}
 \caption{\textbf{Overview of \name}. (A) The proposed framework uses a shared V-Net backbone with two prediction heads for organ segmentation and boundary prediction. Training includes a student--teacher setup with uncertainty-weighted mean-teacher consistency and an axis-flip equivariance loss. (B) Boundary targets are generated automatically from the available organ annotations using 3D erosion and XOR, requiring no additional manual labeling. During inference, only the organ segmentation output is required.}
 \label{fig:pipeline}
 \vspace{-1.5em}
\end{figure*}

Figure~\ref{fig:pipeline} illustrates the architecture of \name. To prevent representation collapse when unlabeled data are scarce ($U \le L$), a shared backbone jointly predicts whole-organ regions alongside auxiliary organ boundaries derived from ground-truth masks. This boundary head serves as an internal anchor to constrain spatial predictions. For unlabeled volumes, uncertainty-gated consistency filters unreliable teacher predictions across both task heads, while axis-flip equivariance reinforces transformation invariance.

\subsection{Architecture \& Boundary Generation}

Let $I \in \mathbb{R}^{H \times W \times D}$ denote an input volumetric scan and $L \in \{0,1\}^{H \times W \times D}$ its corresponding binary segmentation mask. A shared encoder--decoder backbone $\mathcal{F}^{\theta}_{\text{seg}}$, parameterized by weights $\theta$, extracts dense spatial feature representations:
\begin{equation}
    \mathcal{X} = \mathcal{F}^{\theta}_{\text{seg}}(I), \qquad \mathcal{X} \in \mathbb{R}^{C \times H \times W \times D},
\end{equation}
where $C$ denotes the number of feature channels. The representation $\mathcal{X}$ is routed to two task-specific $1\times1\times1$ convolutional heads predicting whole-organ masks ($P_{\text{seg}}$) and organ boundaries ($P_{\text{bd}}$):
\begin{equation}
    P_{\text{seg}} = f_{\theta_1}(\mathcal{X}), \qquad P_{\text{bd}} = f_{\theta_2}(\mathcal{X}).
\end{equation}

The primary segmentation branch is supervised on labeled samples using a combination of Dice loss and Binary Cross-Entropy (BCE) loss:
\begin{equation}
    \mathcal{L}_{\text{Seg}} = \mathcal{L}_{\text{Dice}}(P_{\text{seg}}, L) + \mathcal{L}_{\text{BCE}}(P_{\text{seg}}, L).
\end{equation}

\textbf{Boundary Target Generation.} Ground-truth boundary targets are generated automatically without additional annotation effort. Let $\Gamma_r(L)$ denote the 3D binary erosion of mask $L$ using a spherical structuring element of radius $r$. The boundary target $L_{\text{bd}}$ is computed as the morphological residual:
\begin{equation}
    L_{\text{bd}} = L \oplus \Gamma_r(L),
\end{equation}
where $\oplus$ represents the voxel-wise XOR operation, isolating an internal boundary shell of width $r$.  When training on sub-volume patches, organs truncated by patch borders can introduce artificial crop boundaries. To prevent supervising the network on crop artifacts, boundary voxels within a margin of 2 voxels from patch boundaries are masked out during loss calculation. The boundary prediction is supervised using the same scale-invariant classification loss:
\begin{equation}
    \mathcal{L}_{\text{Boundary}} = \mathcal{L}_{\text{Dice}}(P_{\text{bd}}, L_{\text{bd}}) + \mathcal{L}_{\text{BCE}}(P_{\text{bd}}, L_{\text{bd}}).
\end{equation}

Operating strictly within a single loss space ensures that $\mathcal{L}_{\text{Seg}}$ and $\mathcal{L}_{\text{Boundary}}$ hare identical probability distributions and gradient dynamics. This eliminates the need for loss re-scaling, or post-processing, allowing the boundary head to function as a robust geometric anchor when unlabeled data are limited ($U \le L$).

\subsection{Uncertainty-Weighted Dual-Head Consistency}

To extract reliable supervision from limited unlabeled volumes without triggering confirmation bias, we employ an uncertainty-gated Mean-Teacher framework \cite{tarvainen2017mean,yu2019uncertainty}. The student network ($\theta$) is matched with a teacher network ($\theta'$) updated via exponential moving average (EMA):
\begin{equation}
    \theta' \leftarrow \alpha \theta' + (1-\alpha)\theta,
\end{equation}
where $\alpha$ is the decay momentum. To stabilize early training, the effective momentum $\alpha_t$ ramps up according to $\alpha_t = \min\left(1 - \frac{1}{t+1}, \alpha\right)$ at step $t$.

For an unlabeled volume $I_u$, the student produces raw logits $P_{\text{seg}}$ and $P_{\text{bd}}$, while the teacher generates targets $P'_{\text{seg}}$ and $P'_{\text{bd}}$. Softmax/sigmoid probabilities are denoted as $p_{\text{seg}} = \sigma(P_{\text{seg}})$ and $p'_{\text{seg}} = \sigma(P'_{\text{seg}})$.  Teacher prediction uncertainty at voxel $v$ is estimated using normalized Bernoulli variance:
\begin{equation}
    U_{\text{seg}}(v) = 4 \, p'_{\text{seg}}(v) \left(1 - p'_{\text{seg}}(v)\right) \in [0, 1].
\end{equation}
A binary confidence mask $M_v$ filters out ambiguous voxel predictions where uncertainty exceeds threshold $\tau$:
\begin{equation}
    M_v = \mathbbm{1}\left[U_{\text{seg}}(v) < \tau\right].
\end{equation}

Crucially, this uncertainty mask $M_v$ gates consistency across both organ segmentation and boundary predictions. The dual-head consistency loss $\mathcal{L}_{\text{Cons}}$ is formulated as:
\begin{align*}
    \mathcal{L}_{\text{Cons}} &= \frac{1}{\sum_v M_v} \sum_{v} M_v \Big( \left\| p_{\text{seg}}(v) - p'_{\text{seg}}(v) \right\|_2^2 + \\ 
                            & \qquad \left\| p_{\text{bd}}(v) - p'_{\text{bd}}(v) \right\|_2^2 \Big)
\end{align*}
Note that while $\mathcal{L}_{\text{Cons}}$ uses an $L_2$ metric to enforce smooth consistency, it operates strictly on bounded sigmoid probabilities ($p \in [0, 1]$) across both task heads. This preserves scale symmetry between organ and boundary predictions without introducing the unbounded gradient dynamics typical of continuous distance field regression.

\subsection{Ground-Truth-Anchored Regularization}
To stabilize spatial feature representations when labeled data are scarce, we apply two regularizers anchored strictly to ground-truth annotations, avoiding additional
pseudo-label noise.

\textbf{Axis-Flip Equivariance Regularization.} For spatial flip transformations $\Phi_a(\cdot)$ along anatomical axis $a \in A = \{x, y, z\}$, an equivariant
representation satisfies $f(\Phi_a(I)) \approx \Phi_a(f(I))$. Enforcing this directly as a consistency penalty between $f(\Phi_a(I))$ and $\Phi_a(f(I))$ would require trusting predictions on unlabeled volumes, reintroducing exactly the pseudo-label dependence this section avoids. We therefore induce equivariance \emph{implicitly}, supervising the student on flipped labeled pairs $(\Phi_a(I), \Phi_a(L), \Phi_a(L_{\text{bd}}))$ to yield predictions $\hat{P}_{\text{seg}}^{a}$ and $\hat{P}_{\text{bd}}^{a}$:
\begin{align*}
    \mathcal{L}_{\text{Flip}} &= \frac{1}{|A|} \sum_{a \in A} \Big[ \mathcal{L}_{\text{Seg}}\left(\hat{P}_{\text{seg}}^{a}, \Phi_a(L)\right) + \\
                              & \lambda_b \mathcal{L}_{\text{Boundary}}\left(\hat{P}_{\text{bd}}^{a}, \Phi_a(L_{\text{bd}})\right) \Big].
\end{align*}
Averaging by $1/|A|$ makes the magnitude of $\mathcal{L}_{\text{Flip}}$ independent of how many axes are enforced.

\textbf{Labeled Copy-Paste Augmentation.} To enrich structural variation without
introducing unverified pseudo-labels, copy-paste compositing is restricted strictly
to the labeled stream. With probability $p_{\text{cp}} = 0.5$, a donor labeled pair
$(I_d, L_d)$ is pasted onto a receiver labeled pair $(I, L)$ using foreground mask
$m = \mathbbm{1}[L_d > 0]$:
\begin{equation*}
    \tilde{I} = (1-m) \odot I + m \odot I_d, \qquad \tilde{L} = (1-m) \odot L + m \odot L_d.
\end{equation*}
The boundary target is recomputed from the composited mask as $\tilde{L}_{\text{bd}} = \tilde{L} \oplus \Gamma_r(\tilde{L})$, ensuring anatomical self-consistency across augmented samples. 

\subsection{Overall Training Objective \& Inference}

The overall objective function optimized during training is:
\begin{equation*}
    \mathcal{L}_{\text{Total}} = \mathcal{L}_{\text{Seg}} + \lambda_b \mathcal{L}_{\text{Boundary}} + w(t) \mathcal{L}_{\text{Cons}} + \lambda_f \mathcal{L}_{\text{Flip}},
\end{equation*}
where $\lambda_b$ balances boundary supervision, $\lambda_f$ controls axis-flip regularization, and $w(t) = w_{\max} \cdot \exp\left(-5(1 - t/t_{\max})^2\right)$ is a sigmoid ramp-up function. The ramp-up suppresses consistency loss in early iterations while teacher parameters stabilize. Because $\mathcal{L}_{\text{Flip}}$ replicates the supervised objective on transformed copies, adding it at unit weight would approximately double the supervised gradient; we therefore introduce it with a weight $\lambda_f = 0.5$. 

\section{Results And Discussion}
\textbf{Datasets.} 
We evaluate on four public 3D medical image segmentation benchmarks. \textbf{Left Atrium (LA)} \cite{xiong2021global,bai2023bidirectional} comprises gadolinium-enhanced MRI scans, with 79 training and 20 test cases. \textbf{VerSe} \cite{amiranashvili2024learning} contains CT volumes with vertebra-level labels, split into 190 training and 52 test scans. \textbf{ACDC} \cite{bernard2018deep} provides cine cardiac MRI volumes annotated for three foreground structures and is our only multi-class dataset; we use 160 training and 40 test scans. \textbf{PROMISE12} \cite{litjens2014evaluation} contains T2-weighted prostate MRI volumes, with 50 training and 30 test scans. We vary the number of labeled volumes, treat the remaining cases as unlabeled, and evaluate the final checkpoint without validation.

\textbf{Evaluation Metrics.} We report Dice for volumetric overlap, HD95 for boundary disagreement with reduced outlier sensitivity, and ASD for mean surface distance. Metrics are computed after sliding-window inference at the original resolution.

\noindent\textbf{Baseline Comparisons.} We compare \name against a fully supervised baseline (\textit{FS}) and ten semi-supervised methods: UA-MT \cite{yu2019uncertainty}, DTC \cite{luo2021semi}, SS-Net \cite{wu2022exploring}, BCP \cite{bai2023bidirectional}, CPS \cite{chen2021semi}, MCNet \cite{wu2022mutual}, MCF \cite{wang2023mcf}, CAML \cite{gao2023correlation}, PPC \cite{yuan2024effective}, and UnCo \cite{zeng2025uncertainty}. These methods span consistency-, uncertainty-, geometric-, prototype-, and augmentation-based strategies. To evaluate the low-data setting targeted by this work, all methods receive limited labeled and unlabeled data. This setting is important because many SSL methods assume a large unlabeled pool, which may be unavailable in privacy-constrained clinical applications. Our comparison therefore assesses both label efficiency and robustness to scarce unlabeled data.

\noindent \textbf{Implementation Details.}  All baselines use default configurations with a shared V-Net backbone~\cite{milletari2016v}. Models are trained for $20{,}000$ iterations using SGD with $0.9$ momentum, $10^{-4}$ weight decay, and an initial learning rate of $0.01$, step-decayed every $3{,}333$ iterations. The batch size is $4$ ($2$ labeled and $2$ unlabeled samples for semi-supervised methods). \name{} uses instance normalization due to the small batch size, except on ACDC, where all methods use batch normalization for stability. Hyperparameter tuning sets the erosion kernel size to $r=3$, EMA decay to $\alpha=0.99$, label smoothing to $\epsilon=0.1$, and loss weights to $\lambda_b=1.0$ and $\lambda_f=0.5$. All networks process 3D volumes directly without 2D slice sampling. Input patches are $112^2 \times 80$ (LA), $96^3$ (PROMISE12), $128^3$ (VerSe), and $160^2 \times 16$ (ACDC). LA and PROMISE12 experiments use NVIDIA Titan V GPUs (12,GB), while the higher-memory VerSe and ACDC experiments use NVIDIA A100 GPUs (40,GB). All metrics are averaged over three seeds ($1337$, $42$, and $123$).

\subsection{Results}

\noindent\textbf{Performance Under Unlabeled Data Scarcity for Left Atrium (LA) dataset.} We first evaluate different semi-supervised method sensitivity to unlabeled pool size. Figure~\ref{fig:motivation_figure} sweeps the unlabeled pool from 75 down to 4 volumes under a fixed labeled budget, revealing severe degradation across established paradigms once $U \le L$. At 4 labeled cases, shrinking the unlabeled pool causes UnCo and CPS to drop by 21.4 and 20.5 Dice points, respectively, while CAML, UA-MT, SS-Net, and BCP each lose 10--13 points; only MCF and PPC remain within 2 points of their peak performance. This performance drop highlights a key challenge in low-data scenarios: at a 4:4 ratio, UnCo falls below the strictly supervised baseline, while DTC and CPS provide marginal gains (under 3 Dice points) despite incorporating additional unlabeled volumes. In this context, introducing scarce unlabeled data offers diminishing returns and can occasionally introduce noise that hinders model generalization. This response is notably non-monotonic, exposing the underlying failure mechanism. For CAML, CPS, and SS-Net, a small unlabeled pool yields worse performance than virtually none: at 8 labeled cases, expanding the unlabeled pool from 4 to 8 volumes decreases CAML's performance by 3.7 Dice points and CPS's by 3.2 points before both recover at larger pool sizes. Consistency and pseudo-labeling objectives rely on averaging teacher prediction errors across diverse unlabeled samples. When $U \le L$, the lack of sample diversity prevents error cancellation; instead, unconfident or noisy targets reinforce confirmation bias. Standard SSL benchmarks obscure this vulnerability by fixing the unlabeled pool to the full dataset and varying only the labeled count. Figure~\ref{fig:motivation_figure} provides a systematic characterization of this failure mode in 3D semi-supervised medical image segmentation. 

\noindent\textbf{\name is less-sensitive to unlabeled data Size.} Across the sweep, \name maintains stable performance, varying by only 0.40 and 0.46 Dice points at 4 and 8 labeled cases, respectively (seed variance $<1.2$). This structural stability stems from deriving boundary targets deterministically from ground-truth masks via morphological erosion and XOR, maintaining anchor strength regardless of unlabeled pool size. Consequently, \name achieves its largest performance margins with smallest unlabeled data, whereas pool-dependent baselines like BCP require large cohorts to peak ($88.4$ vs.\ $86.8$ Dice at 4:75). However, as the unlabeled pool expands, competing methods become increasingly effective; notably, BCP outperforms \name under the larger 4:75 and is tied at 8:71 splits (\Cref{tab:la_supp}).

\noindent\textbf{Head-to-Head Performance at $U \le L$.} Table~\ref{tab:la_results_vertical} reports quantitative performance under strict data constraints ($U \le L$), fixing the unlabeled pool at 4 volumes across two labeled budgets. At 4:4 ($U=L$), \name achieves 86.95 Dice, outperforming the strongest baseline (78.20 Dice) by 8.75 points while achieving superior HD95 and ASD surface metrics. At 8:4 ($U<L$), the Dice margin narrows to 2.75 points, and surface distance metrics converge, with our HD95 reaching statistical parity with MCF ($p = 0.57$, paired Wilcoxon signed-rank test across 20 test cases). This narrowing aligns with our theoretical framing: expanding the labeled set provides explicit structural supervision that partially substitutes for the spatial regularizing role of our boundary objective. Consequently, \name provides the largest performance gains under the most severely constrained supervision scenarios.

Seed sensitivity separates the methods almost as clearly as the means do. At 4:4, DTC varies by $\pm7.7$ Dice across three seeds and CAML by $\pm4.9$, while UnCo's HD95 spread of $\pm12.4$mm is larger than the entire margin most methods hold over the labeled-only reference---a single seed is not representative of these methods' behaviour on this split. \name is the most stable entry at $\pm1.2$ Dice, with SS-Net closest at $\pm1.4$. At 8:4 the field tightens considerably, and the ordering changes: \name ($\pm0.47$), BCP ($\pm0.46$) and MCNet ($\pm0.45$) are effectively tied for the lowest variance, while DTC ($\pm5.2$) and CPS ($\pm3.4$) remain unstable. Methods exhibiting high variance across both budgets—namely DTC, CPS, and UnCo (on surface metrics)—are the exact methods that fail to beat the labeled-only baseline. This overlap points to a single root cause: an unreliable training signal on a dataset too small to stabilize it.

\begin{figure*}
    \centering
    \includegraphics[width=0.9\linewidth]{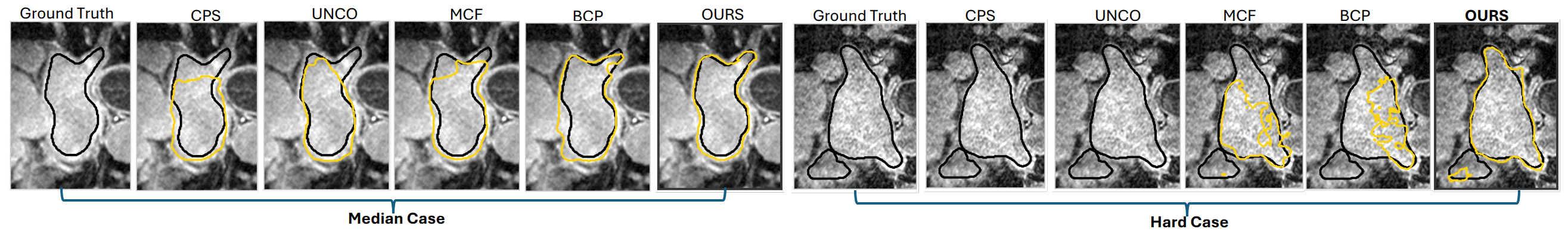}
     \caption{\textbf{Qualitative results on LA (4 labeled / 4 unlabeled).} Hardest and median test cases, selected by mean Dice across all six methods rather than by our own. Same axial slice per row; black: ground truth, yellow: prediction. On the hardest case three baselines fail entirely ($\le0.001$) and two fragment ($0.671$, $0.342$), while \name returns a single coherent boundary.}
    \label{fig:la_qualitative}
    \vspace{-1.5em}
\end{figure*}

\noindent\textbf{Qualitative Comparison.} Figure~\ref{fig:la_qualitative} illustrates segmentation performance across representative test cases. On the most challenging test volume (selected by average mean dice across all methods), three of five competing methods fail to segment the foreground, while the remaining two produce fragmented predictions characterized by internal cavities and spurious components. In contrast, \name yields a single closed contour that correctly delineates the atrial wall, preserving fine anatomical structures such as the inferior extension missed by all baselines. Median cases exhibit similar topological degradation: while baselines capture the primary atrial body, they frequently truncate the thin, high-curvature superior pulmonary-vein extensions. By preserving topological integrity and continuous boundaries, \name achieves pronounced improvements in surface distance metrics (HD95 and ASD) relative to standard region-only architectures.

\begin{table}[t]
\centering
\small
\setlength{\tabcolsep}{4pt}
\caption{\textbf{LA results}. \textit{Labeled} is labelled-only baseline. For the 8:4 split, the HD95 scores of MCF and our method are statistically indistinguishable under a paired Wilcoxon signed rank test. Bold indicates the best statistically significant method under a paired $t$-test, two bold means statistical tie with $p>0.05$.}
\label{tab:la_results_vertical}
{%
\begin{tabular}{llccc}
\toprule
Split & Method & Dice & HD95 & ASD \\
\midrule
\multirow{12}{*}{4:4}
& \textit{Labeled}       & 58.77$\pm$5.44 & 36.17$\pm$3.34 & 4.96$\pm$1.54 \\
& UA-MT    & 69.35$\pm$3.16 & 26.24$\pm$2.65 & 6.43$\pm$1.53 \\
& DTC      & 61.03$\pm$7.68 & 36.05$\pm$8.50 & 6.57$\pm$2.43 \\
& SS-Net   & 73.45$\pm$1.40 & 22.82$\pm$2.70 & 3.40$\pm$0.25 \\
& BCP      & 78.20$\pm$2.48 & 14.33$\pm$2.48 & 3.32$\pm$0.31 \\
& CPS      & 61.61$\pm$3.74 & 34.72$\pm$1.68 & 8.90$\pm$3.05 \\
& MCNet    & 73.03$\pm$3.73 & 24.61$\pm$4.61 & 4.03$\pm$1.17 \\
& MCF      & 78.06$\pm$2.83 & 16.39$\pm$3.10 & 5.02$\pm$1.26 \\
& CAML     & 70.26$\pm$4.91 & 28.07$\pm$5.68 & 3.72$\pm$0.63 \\
& PPC      & 75.19$\pm$3.48 & 17.50$\pm$2.00 & 4.50$\pm$0.81 \\
& UnCo     & 57.01$\pm$2.97 & 38.72$\pm$12.41 & 4.30$\pm$0.73 \\
\rowcolor{gray!35}
& \textbf{Ours} & \textbf{86.95$\pm$1.21} & \textbf{10.25$\pm$2.61} & \textbf{2.87$\pm$1.13} \\
\midrule
\midrule
\multirow{12}{*}{8:4}
& \textit{Labeled}   & 81.11$\pm$0.75 & 14.97$\pm$2.22 & 2.61$\pm$0.10 \\
& UA-MT    & 83.63$\pm$1.43 & 12.70$\pm$1.10 & 2.88$\pm$0.42 \\
& DTC      & 76.60$\pm$5.18 & 19.29$\pm$3.11 & 2.92$\pm$0.60 \\
& SS-Net   & 85.72$\pm$1.06 & 10.71$\pm$0.90 & 2.25$\pm$0.15 \\
& BCP      & 81.71$\pm$0.46 & 13.86$\pm$0.76 & 2.50$\pm$0.41 \\
& CPS      & 80.27$\pm$3.44 & 16.70$\pm$2.11 & 3.45$\pm$0.27 \\
& MCNet    & 86.54$\pm$0.45 & 11.20$\pm$0.79 & \textbf{2.03$\pm$0.08} \\
& MCF      & 85.20$\pm$0.92 & \textbf{8.05$\pm$0.44} & 2.35$\pm$0.09 \\
& CAML     & 83.44$\pm$2.81 & 15.50$\pm$3.09 & \bf 2.07$\pm$0.08 \\
& PPC      & 84.21$\pm$0.88 & 11.27$\pm$0.57 & 2.50$\pm$0.10 \\
& UnCo     & 79.68$\pm$2.37 & 16.90$\pm$1.11 & 2.68$\pm$0.33 \\
\rowcolor{gray!35}
& \textbf{Ours} & \textbf{89.29$\pm$0.47} & \textbf{8.58$\pm$1.23} & 2.26$\pm$0.43  \\
\bottomrule
\end{tabular}%
}
\vspace{-1.0em}
\end{table}

\noindent\textbf{Results on VerSe.} VerSe tests whether \name's behaviour is specific to cardiac MRI (LA), with results reported in Table~\ref{tab:verse_results} and \cref{tab:verse_results_55_single}. At 5:5, nine of ten baselines fall below the labeled-only reference of $62.2$ Dice. UnCo is the sole exception at $64.8$, but its seed spread ($\pm9.4$) is more than three times its $2.6$-point gain. BCP ($61.3$) and PPC ($58.7$) come closest without clearing the reference, while CPS ($19.4$) and DTC ($20.6$) collapse to a third of it and CAML ($30.1$) and MCNet ($30.9$) to half. Whereas LA showed that scarce unlabeled data erodes the benefit of semi-supervision, VerSe reverses its sign for all but one method. Doubling the labeled budget only partly repairs this: at 10:5, SS-Net, BCP, PPC, and UnCo clear the reference, but CPS still trails it by $23$ points and CAML by $44$. Instability is also evident in the seed spreads---CAML varies by $\pm16.9$ at 5:5 and $\pm23.9$ at 10:5, BCP by $\pm23.3$, and DTC by $\pm13.8$, on the order of their means. We attribute this to the multi-instance task: adjacent vertebrae are near-identical, allowing a mis-localised teacher to produce targets that are locally plausible but globally wrong, while five unlabeled volumes are too few for such errors to cancel.

\name clears the reference at both splits by $21.6$ and $2.9$ Dice ($83.7$ and $89.0$), with the table's lowest variance ($\pm0.78$ and $\pm0.27$). UnCo is the only other method above the reference at both, by $2.6$ and $0.1$ points against spreads of $\pm9.4$ and $\pm1.4$. At 5:5, \name leads Dice and HD95 significantly against every baseline ($p<0.001$, paired Wilcoxon over $52$ cases) and matches BCP on ASD ($p{=}0.20$). At 10:5, the Dice lead remains ($p{=}0.003$ vs.\ BCP), but surface metrics no longer separate the leading methods: SS-Net ($4.93$ HD95, $1.40$ ASD) and UnCo ($5.29$, $1.54$) are nominally ahead of \name ($5.73$, $1.79$), with SS-Net and BCP previously tested as ties ($p{=}0.46$, $p{=}0.40$). With more labels, the advantage therefore lies in overlap rather than surface quality---the same narrowing-with-labels pattern observed on LA, but from a much lower baseline. Qualitative examples are provided in Fig.~\ref{fig:verse_qualitative}. On the hardest case, baselines either nearly miss the vertebra ($0.02$--$0.05$ Dice) or scatter fragments into surrounding tissue, whereas \name recovers a coherent vertebral body ($0.66$). On the median case, their dominant error is leakage into the vertebral foramen and adjacent processes.
\begin{table}[!thb]
\centering
\scriptsize
\setlength{\tabcolsep}{4pt}
\caption{\textbf{VerSe results for the 5:5 split}: \textit{Labeled} is model
trained only on labeled data. SGTC has no VerSe runs and is omitted.}
\label{tab:verse_results_55_single}
\resizebox{\columnwidth}{!}{%
\begin{tabular}{lcccc}
\toprule
 Method & Dice & Jaccard & HD95 & ASD \\
\midrule
\textit{Labeled}& 62.17$\pm$11.50 & 49.16$\pm$12.28 & 12.49$\pm$2.35 & 3.79$\pm$0.74 \\
 UA-MT            & 34.33$\pm$8.70  & 24.62$\pm$8.12  & 32.75$\pm$5.34 & 16.27$\pm$3.22 \\
 DTC              & 20.60$\pm$13.84 & 14.01$\pm$10.78 & 39.18$\pm$12.58 & 8.55$\pm$3.94 \\
 SS-Net           & 49.71$\pm$9.15  & 35.91$\pm$8.07  & 19.52$\pm$9.63 & 7.82$\pm$5.53 \\
 BCP              & 61.34$\pm$23.30 & 49.94$\pm$25.00 & 11.73$\pm$6.28 & \textbf{2.54$\pm$0.65} \\
 CPS              & 19.36$\pm$3.85  & 11.74$\pm$2.37  & 45.18$\pm$6.65 & 24.99$\pm$5.02 \\
 MCNet            & 30.89$\pm$10.03 & 22.23$\pm$9.35  & 33.92$\pm$5.13 & 13.58$\pm$5.48 \\
 MCF              & 53.42$\pm$11.13 & 38.78$\pm$10.92 & 21.23$\pm$5.79 & 8.13$\pm$2.37 \\
 CAML             & 30.12$\pm$16.85 & 21.52$\pm$12.73 & 32.23$\pm$10.32 & 6.99$\pm$3.12 \\
 PPC              & 58.73$\pm$6.99  & 43.10$\pm$7.60  & 20.95$\pm$4.67 & 7.00$\pm$1.61 \\
 UnCo             & 64.77$\pm$9.41  & 50.71$\pm$9.48  & 12.63$\pm$3.67 & 4.15$\pm$1.66 \\
\rowcolor{gray!35}
 \textbf{Ours}    & \textbf{83.73$\pm$0.78} & \textbf{74.40$\pm$1.36} & \textbf{8.16$\pm$0.81} & \bf  2.55$\pm$0.26 \\
\bottomrule
\end{tabular}%
}
\vspace{-1.5em}
\end{table}

\noindent\textbf{Results on ACDC.} On the multi-class ACDC benchmark(\cref{tab:acdc_results}) (right ventricle, myocardium, and left ventricle), standard SSL paradigms degrade markedly under strict data constraints. At 14:14 ($U=L$), all six baselines perform at or below the supervised reference ($68.4$ Dice): BCP is comparable at $67.6$ and PPC reaches $61.3$, whereas MCF ($53.5$), UA-MT ($49.3$), CAML ($30.5$), and MC-Net ($28.1$) underperform supervised training. At 7:14, only BCP clearly surpasses the reference ($59.5$ vs.\ $48.3$); PPC exceeds it by just $0.7$ Dice against a $\pm5.7$ seed spread, and no other method approaches it. Multi-class targets compound error propagation because noisy teacher predictions introduce both categorical errors and spatial drift, causing MC-Net and CAML to collapse, with HD95 errors of $137$--$183$,mm across both splits. In contrast, \name leads both metrics and splits, reaching $67.6$ Dice at 7:14 (+19.3 over supervised, +8.1 over BCP) and $77.8$ at 14:14 (+9.4 and +10.2), with all improvements significant ($p<10^{-4}$). The boundary advantage is clearest at 14:14, where \name reduces HD95 to $4.5$,mm---under one-third of the best baseline ($16.2$,mm). At 7:14, the margin over BCP remains significant ($18.9$ vs.\ $20.5$,mm, $p=0.019$). Qualitative results appear in \Cref{fig:acdc_qualitative}.

    \begin{table}[!htb]
    \centering
    \scriptsize
    \setlength{\tabcolsep}{3pt}
    \caption{\textbf{ACDC results.} \textit{Labeled} is trained on the split's
    labeled cases only. UnCo has no ACDC runs and is omitted.}
    \label{tab:acdc_results}
    \resizebox{\columnwidth}{!}{%
    \begin{tabular}{lcc||cc}
    \toprule
    \multirow{2}{*}{Method} & \multicolumn{2}{c||}{7:14} & \multicolumn{2}{c}{14:14} \\
    \cmidrule(lr){2-3}\cmidrule(lr){4-5}
    & Dice & HD95 & Dice & HD95 \\
    \midrule
    \textit{Labeled} & 48.26$\pm$0.95 & 56.70$\pm$5.00  & 68.43$\pm$3.18 & 14.60$\pm$7.51 \\
    UA-MT            & 32.10$\pm$4.37 & 56.90$\pm$13.84 & 49.32$\pm$5.83 & 44.16$\pm$6.62 \\
    BCP              & 59.48$\pm$7.25 & 20.49$\pm$7.24  & 67.61$\pm$2.54 & 16.23$\pm$3.59 \\
    MCNet            & 9.53$\pm$6.47  & 182.82$\pm$26.50 & 28.05$\pm$2.03 & 150.76$\pm$9.88 \\
    MCF              & 36.10$\pm$3.11 & 52.77$\pm$6.82  & 53.53$\pm$1.99 & 31.84$\pm$3.34 \\
    CAML             & 17.23$\pm$11.75 & 181.75$\pm$69.05 & 30.54$\pm$13.49 & 137.51$\pm$50.82 \\
    PPC              & 48.99$\pm$5.70 & 40.60$\pm$10.93 & 61.28$\pm$1.72 & 31.57$\pm$10.96 \\
    \rowcolor{gray!35}
    \textbf{Ours}    & \textbf{67.58$\pm$3.75} & \textbf{18.86$\pm$6.87} & \textbf{77.83$\pm$0.91} & \textbf{4.53$\pm$0.59} \\
    \bottomrule
    \end{tabular}%
    }
    \vspace{-1.5em}
    \end{table}

  \noindent\textbf{Results on Promise12.}At a 5:5 split shown in \cref{tab:promise12_results}, eight of the ten baselines perform at or below the supervised reference ($48.3$ Dice), with PPC ($8.1$) and UnCo ($9.2$) suffering near-complete prediction collapse. Only BCP ($62.2$) and MCF ($60.5$) extract effective supervisory signals from the five unlabeled volumes. This instability is further reflected in high seed-to-seed variance: CPS and UA-MT exhibit spreads of $\pm26.2$ and $\pm22.2$ Dice points across three seeds—variances comparable to their mean performance—underscoring the optimization fragility of standard SSL paradigms in low-data regimes. In contrast, \name achieves $71.5$ Dice (+23.3 over the supervised baseline, +9.4 over the strongest competing method), maintaining statistically significant margins against every baseline ($p \le 0.002$). On the small, highly variable prostate volumes of PROMISE12, surface distance metrics (HD95 and ASD) do not statistically differentiate among successful methods, with \name performing comparably to BCP, MCF, MC-Net, and CAML. Consequently, the performance gain on this dataset primarily demonstrates consistent organ localization and structural recovery rather than localized boundary refinement.

\subsection{Ablation Experiments}

All ablation studies are conducted on the Left Atrium (LA) dataset on the same GPU. 

\noindent\textbf{Erosion kernel-size ablation on LA.}
Table~\ref{tab:kernel_ablation} examines sensitivity to the erosion kernel size $r$, which controls the thickness of the boundary shell extracted by $L \oplus \Gamma_r(L)$. Smaller kernels concentrate supervision near the organ surface, whereas larger kernels extend it farther into the interior. Overall, \name is robust to this choice. Across $r\in{3,5,7}$, mean Dice varies by only 0.8 points at 4:4 and 0.3 points at 8:4, comparable to seed variability. No size consistently performs best: $r=3$ leads at 4:4, whereas $r=5$ leads at 8:4 and has the lowest variance ($\pm0.03$ Dice). This indicates that the benefit primarily arises from explicitly supervising the boundary region rather than selecting a particular shell thickness. We choose $r=3$ in the most annotation-constrained 4:4 setting and fix it across all label budgets and datasets, avoiding split-specific tuning.

 \textbf{Component ablation.} Table~\ref{tab:la_ablation} adds one component at a time at the most label-scarce setting (4:4), recovering $+30.46$ Dice and $-29.2$mm HD95 over the supervised floor. The boundary head is the largest single contribution ($+9.91$ Dice, $p<10^{-4}$), followed by normalization ($+6.32$) and mean-teacher consistency ($+7.90$); seed variance contracts  from $\pm6.71$ to $\pm1.21$ Dice as components accumulate, so the additions stabilise training as well as improve it.

\textbf{Classification vs.\ Regression Supervision.} To verify that the classification-space formulation provides the core benefit, rather than merely adding an auxiliary head, we replace the boundary head with an SDF regression head under an identical setup. Classification significantly outperforms regression, improving Dice by $1.40$, HD95 by $2.34$,mm, and ASD by $0.69$,mm, with the largest gains in surface metrics. Sweeping the auxiliary weight over two orders of magnitude does not close this gap ($11/12$ comparisons favor classification, $p<0.05$). Because the targets differ in representation and spatial extent, we also supervise the \emph{same} distance information, discretised into ordinal bands and trained with the boundary head's classification loss. This matches the boundary head ($p{=}0.78$), whereas regressing the same information loses $1.76$ Dice. The advantage therefore arises from the loss formulation, not the target's geometric content. Full results are provided in Supplementary Table~\ref{tab:aux_head_swap}.

\textbf{Flip Axes Ablation.} Equivariance across all three anatomical axes is essential: using one or two axes recovers at most half of the Dice improvement ($\le 86.21\%$ vs.\ $86.95\%$), while HD95 remains stagnant ($12.2\text{--}13.8$~mm) until dropping sharply to $10.25$~mm at $k=3$. Because flips are applied solely to the labeled stream, computational overhead is minimal: enforcing three axes takes only $1.26\times$ the per-iteration time of a single axis, with the second axis adding just $5\%$ (Appendix~\ref{tab:flip_axes}). We therefore adopt $k=3$ in all experiments.

\section{Conclusion and Future Work}

We presented \name, a multi-task semi-supervised framework designed to stabilize 3D medical image segmentation in the data-constrained regime ($U \le L$). By deriving auxiliary boundary targets deterministically from ground-truth masks via morphological erosion and XOR, \name provides an internal structural anchor that prevents pseudo-label degeneracy without relying on large unlabeled cohorts. Integrated with uncertainty-weighted dual-head consistency, axis-flip equivariance, and copy-paste regularization, \name consistently outperforms competitive semi-supervised baselines across four public 3D benchmarks when unlabeled data are limited. Crucially, because auxiliary heads and teacher networks are discarded post-training, the framework incurs zero additional inference overhead.

\textit{Limitations and future work}. Despite these strengths, performance on PROMISE12 highlights a key limitation: while \name significantly improves volumetric overlap, its gains on surface distance metrics (HD95 and ASD) are less pronounced. This indicates that fixed binary targets may lack precision for small, highly variable structures with diffuse margins. Future work will explore adaptive boundary formulations that dynamically scale the erosion radius $r$ based on local anatomical volume and label scarcity. Additionally, we plan to validate \name on multi-center, multi-modal clinical cohorts to assess its robustness against domain shift in low-data regimes.

\clearpage
{
    \small
    \bibliographystyle{ieeenat_fullname}
    \bibliography{main}
}
\clearpage
\section{Supplementary Section}
\setcounter{section}{0}
\setcounter{figure}{0}
\setcounter{table}{0}
\renewcommand{\thesection}{S\arabic{section}}
\renewcommand{\thefigure}{S\arabic{figure}}
\renewcommand{\thetable}{S\arabic{table}}
\renewcommand{\theequation}{S\arabic{equation}}

\subsection{Additional Results and Analysis}

\begin{table}[!htb]
\centering
\small
\setlength{\tabcolsep}{4pt}
\caption{\textbf{Promise12 results at 5 labeled / 5 unlabeled cases}.
\textit{Labeled} is trained on the labeled cases only. Bold marks the best statistically significant method on Dice (paired Wilcoxon over 30 test cases, $p\le0.002$ against every baseline). No bolding is applied to HD95 or ASD: apart from the four methods that fail outright, the surface metrics are not significantly separated on this dataset.}
\label{tab:promise12_results}
\begin{tabular}{lccc}
\toprule
Method & Dice & HD95 & ASD \\
\midrule
\textit{Labeled} & 48.26$\pm$3.77 & 19.18$\pm$6.77 & 8.07$\pm$4.55 \\
UA-MT   & 43.71$\pm$22.18 & 37.03$\pm$30.48 & 21.27$\pm$19.12 \\
DTC     & 46.52$\pm$6.29  & 26.75$\pm$10.30 & 13.93$\pm$7.19 \\
SS-Net  & 41.48$\pm$8.93  & 23.51$\pm$5.81  & 10.78$\pm$6.00 \\
BCP     & 62.18$\pm$12.48 & 22.01$\pm$10.84 & 9.79$\pm$5.40 \\
CPS     & 44.63$\pm$26.21 & 39.30$\pm$34.70 & 22.56$\pm$21.59 \\
MCNet   & 48.16$\pm$17.40 & 26.16$\pm$3.51  & 14.42$\pm$0.89 \\
MCF     & 60.52$\pm$12.90 & 18.46$\pm$7.60  & 9.31$\pm$4.78 \\
CAML    & 43.60$\pm$13.17 & 17.22$\pm$7.98  & 8.13$\pm$3.74 \\
PPC     & 8.08$\pm$4.48   & 71.18$\pm$6.06  & 50.38$\pm$7.01 \\
UnCo    & 9.16$\pm$6.73   & 60.33$\pm$9.62  & 38.82$\pm$11.28 \\
\rowcolor{gray!35}
\textbf{Ours} & \textbf{71.53$\pm$9.15} & 18.16$\pm$10.42 & 11.08$\pm$9.32 \\
\bottomrule
\end{tabular}
\end{table}

\begin{table}[!htb]
\centering
\caption{\textbf{VerSe results.} Columns are labeled:unlabeled case counts.
\textbf{Bold} marks the best value in each column among the semi-supervised
methods. \textit{Labeled} is trained on the split's labeled cases only and is
excluded from bolding; at 10:5, only four of ten semi-supervised methods exceed
it in Dice.}
\label{tab:verse_results}
{%
\begin{tabular}{lccc}
\toprule
Method & Dice & HD95 & ASD \\
\midrule
\textit{Labeled} & 86.07$\pm$1.44 & 5.17$\pm$0.91 & 1.62$\pm$0.36 \\
\midrule
UA-MT  & 80.99$\pm$1.20 & 9.14$\pm$0.68 & 3.96$\pm$0.43 \\
DTC    & 82.51$\pm$3.65 & 7.25$\pm$1.76 & 2.63$\pm$0.97 \\
SS-Net & 86.56$\pm$0.66 & \textbf{4.93$\pm$0.46} & \textbf{1.40$\pm$0.10} \\
BCP    & 87.09$\pm$3.25 & 5.97$\pm$2.59 & 1.48$\pm$0.64 \\
CPS    & 62.84$\pm$1.30 & 18.57$\pm$2.19 & 6.86$\pm$0.80 \\
MCNet  & 85.29$\pm$3.03 & 7.01$\pm$2.56 & 2.12$\pm$0.83 \\
MCF    & 79.18$\pm$3.38 & 8.95$\pm$0.31 & 3.02$\pm$0.19 \\
CAML   & 42.32$\pm$23.90 & 28.11$\pm$14.61 & 5.13$\pm$3.35 \\
PPC    & 86.47$\pm$1.70 & 5.91$\pm$0.85 & 1.81$\pm$0.19 \\
UnCo   & 86.15$\pm$1.40 & 5.29$\pm$0.62 & 1.54$\pm$0.26 \\
\midrule
\rowcolor{gray!35}
\textbf{Ours} & \textbf{88.98$\pm$0.27} & 5.73$\pm$0.29 & 1.79$\pm$0.10 \\
\bottomrule
\end{tabular}%
}
\end{table}

\begin{table}[!htb]
\centering
\scriptsize
\setlength{\tabcolsep}{3pt}
\caption{\textbf{Erosion kernel size ablation on LA.} The boundary target is the residual $L \oplus \Gamma_r(L)$; $r$ controls the thickness of the extracted surface shell.}
\label{tab:kernel_ablation}
\resizebox{\columnwidth}{!}{%
\begin{tabular}{lccc|ccc}
\toprule
\multirow{2}{*}{$r$} & \multicolumn{3}{c|}{4:4} & \multicolumn{3}{c}{8:4} \\
\cmidrule(lr){2-4}\cmidrule(lr){5-7}
& Dice & HD95 & ASD & Dice & HD95 & ASD \\
\midrule
3 & \textbf{86.95$\pm$1.21} & \textbf{10.25$\pm$2.61} & \textbf{2.87$\pm$1.13} & 89.29$\pm$0.47 & 8.58$\pm$1.23 & 2.26$\pm$0.43 \\
5 & 86.19$\pm$1.21 & 12.34$\pm$1.58 & 3.57$\pm$0.72 & \textbf{89.50$\pm$0.03} & \textbf{7.15$\pm$0.29} & \textbf{1.83$\pm$0.03} \\
7 & 86.32$\pm$0.33 & 12.17$\pm$0.70 & 3.41$\pm$0.20 & 89.17$\pm$0.53 & 8.82$\pm$1.47 & 2.36$\pm$0.54 \\
\bottomrule
\end{tabular}%
}
\end{table}

\noindent\textbf{Training Computational Overhead.} Table~\ref{tab:la_cost} compares the computational footprint of \name against representative SSL baselines under identical Left Atrium (LA) settings. \name exhibits a training-time profile comparable to standard single-teacher paradigms, requiring 18.89\,M parameters and 0.839\,s/iter—matching UA-MT and SS-Net while slightly exceeding CPS and MCF. Compared with parameter-heavy architectures or complex dual-network frameworks like UnCo and CAML, \name is substantially lighter and more memory-efficient. Its peak memory footprint (8.46\,GB) remains moderate, well within typical GPU limits. Crucially, because the auxiliary boundary head and EMA teacher network are discarded post-training, all evaluated methods deploy as an identical single V-Net backbone (9.44\,M parameters); thus, the added complexity of \name is strictly confined to training-time regularization.

\begin{table*}[!htb]
\centering
\caption{\textbf{Training overhead comparison on LA.} Measured on an NVIDIA A100 (40\,GB) GPU under identical batch settings (8 labeled cases, patch size $112\times112\times80$, total batch size 4 / 2 labeled). Parameter counts reflect all active components during training (EMA teacher, auxiliary heads); at inference, all methods reduce to a single V-Net backbone (9.44\,M parameters). BCP uses a batch size of 8 / 4 labeled due to its internal batch-halving mechanism, processing 4 volumes per step.}
\label{tab:la_cost}
\begin{tabular}{lrrcc}
\toprule
Method & Time (s/iter) & vs.\ Ours & Train Params (M) & Peak Mem.\ (GB) \\
\midrule
UA-MT  & 0.895 & 1.07$\times$ & 18.90 & 6.53 \\
MCF    & 0.719 & 0.86$\times$ & 27.38 & 7.89 \\
CAML   & 1.849 & 2.20$\times$ & 20.13 & 13.38 \\
PPC    & 1.591 & 1.90$\times$ & 18.94 & 20.05 \\
UnCo   & 0.852 & 1.02$\times$ & 37.80 & 13.90 \\
CPS    & 0.633 & 0.75$\times$ & 18.90 & 6.96 \\
SS-Net & 0.873 & 1.04$\times$ & 9.46  & 10.79 \\
BCP    & 1.607 & 1.92$\times$ & 18.91 & 3.72 \\
\rowcolor{gray!35}
\textbf{\name (Ours)} & 0.839 & 1.00$\times$ & 18.89 & 8.46 \\
\bottomrule
\end{tabular}
\end{table*}

\noindent\textbf{Classification vs.\ Regression Boundary Formulations.}
Table~\ref{tab:aux_head_swap} evaluates the central formulation choice of \name by comparing binary boundary classification (Dice + Cross-Entropy) with continuous Signed Distance Field (SDF) regression ($L_1$ loss), while keeping all other components fixed. At matched auxiliary-loss weights ($\lambda_b=1.0$), classification improves Dice by 1.40 points and boundary accuracy substantially, reducing HD95 by 2.34\,mm and ASD by 0.69\,mm.

We next examine whether this gap is caused by loss weighting. Varying the SDF regression weight across two orders of magnitude ($\lambda_b\in\{0.1,1.0,10.0\}$) changes Dice by only 0.70 points, and no setting approaches the classification head. Because binary boundaries and continuous SDFs differ in both target representation and spatial extent, we additionally evaluate \emph{ordinal distance bands}. These bands preserve discretized distance information through $K{=}3$ nested shells but are optimized using the same Dice + Cross-Entropy objective as the boundary head. The band and boundary heads perform comparably (87.31 vs.\ 86.95 Dice; $p{=}0.78$), whereas SDF regression at the same weight underperforms the band formulation by 1.76 Dice points. Matching the auxiliary gradient magnitude also fails to close the gap: weighting the SDF loss so that its gradient norm on the shared trunk matches that of the boundary head ($\lambda_b=3.0$) produces the lowest Dice score (84.49). Together, these results indicate that the classification-based formulation, rather than simply the removal of distance information or insufficient auxiliary supervision, is the primary source of the improvement.

The same pattern holds at the larger label budget. At 8:4, the boundary and band heads again achieve comparable Dice (89.29 vs.\ 89.82; $p{=}0.93$), with bands nominally improving all three metrics. The band formulation is also substantially more reproducible across both splits: its seed-level standard deviations are $2.5$--$4\times$ smaller than those of the boundary head in all six metric comparisons (e.g., Dice $\pm1.21\rightarrow\pm0.30$ at 4:4 and $\pm0.47\rightarrow\pm0.16$ at 8:4). This suggests that discretized distance supervision can stabilize optimization while yielding similar overall performance. We retain the binary boundary head in \name for simplicity and use the band results to show that the advantage over SDF regression does not arise merely from discarding distance information.

\begin{table*}[!htb]
\centering
\small
\setlength{\tabcolsep}{5pt}
\caption{\textbf{Comparison of auxiliary-task formulations} on LA (mean~$\pm$~std over 3 seeds). All components are identical except for the auxiliary target and loss. Ordinal bands retain discretized distance information from the SDF while casting supervision as classification. The SDF weight sweep was conducted only for the 4:4 split. $^*$ denotes $p<0.05$ relative to the boundary head using a paired Wilcoxon signed-rank test over 20 test cases.}
\label{tab:aux_head_swap}
\begin{tabular}{llccc}
\toprule
Auxiliary Task & $\lambda_b$ & Dice~$\uparrow$ & HD95 (mm)~$\downarrow$ & ASD (mm)~$\downarrow$ \\
\midrule
\multicolumn{5}{l}{\emph{4 labeled / 4 unlabeled}} \\
\rowcolor{gray!25}
Boundary (Classification) & 1.0 & 86.95$\pm$1.21 & 10.25$\pm$2.61 & 2.87$\pm$1.13 \\
Ordinal bands (Classification, $K{=}3$) & 1.0 & \bf 87.31$\pm$0.30 & \bf 9.91$\pm$1.05 & \bf 2.75$\pm$0.34 \\
\multirow{4}{*}{SDF (Regression)}
 & 0.1  & 85.39$\pm$0.58 & 12.97$\pm$0.53$^*$ & 3.77$\pm$0.06$^*$ \\
 & 1.0  & 85.55$\pm$0.39$^*$ & 12.59$\pm$0.35 & 3.56$\pm$0.07$^*$ \\
 & 3.0  & 84.49$\pm$1.48$^*$ & 13.48$\pm$2.23$^*$ & 4.06$\pm$0.79$^*$ \\
 & 10.0 & 84.85$\pm$0.86$^*$ & 14.28$\pm$0.88$^*$ & 4.14$\pm$0.48$^*$ \\
\midrule
\multicolumn{5}{l}{\emph{8 labeled / 4 unlabeled}} \\
\rowcolor{gray!25}
Boundary (Classification) & 1.0 & 89.29$\pm$0.47 & 8.58$\pm$1.23 & 2.26$\pm$0.43 \\
Ordinal bands (Classification, $K{=}3$) & 1.0 & \bf 89.82$\pm$0.16 & \bf 7.52$\pm$0.48 & \bf 1.92$\pm$0.16 \\
\bottomrule
\end{tabular}
\end{table*}

\noindent\textbf{Impact of Transformation Equivariance Axes.} Table~\ref{tab:flip_axes} ablates axis-flip equivariance across anatomical dimensions ($D, H, W$). Performance does not scale monotonically with the number of axes $k$. Single-axis symmetry yields marginal improvements over the baseline ($85.16 \rightarrow 85.78$ for axis $H$), while enforcing equivariance exclusively along the axial dimension ($D$) slightly degrades performance ($85.08$). At $k=2$, performance depends heavily on axis selection: combining $D+H$ yields $85.03$ Dice, whereas $H+W$ reaches $86.21$. Omitting the axial dimension in partial subsets degrades representation stability; full three-axis equivariance ($k=3$) is required to unlock maximum gains, outperforming the best dual-axis pair by $+0.74$ Dice.

Surface metrics demonstrate an even stronger dependence on complete axial coverage: HD95 remains virtually unchanged across all single- and dual-axis configurations ($12.20$--$13.82$\,mm) before dropping sharply to $10.25$\,mm when $k=3$. This indicates that spatial representation regularization requires removing rotational ambiguity across all three dimensions simultaneously. Computational overhead for full equivariance remains modest: applying transformations exclusively to the labeled stream increases training time from 0.96\,s/iter ($k=0$) to 1.49\,s/iter ($k=3$), representing a $1.26\times$ runtime increase relative to single-axis enforcement.

\begin{table}[!htb]
\centering
\small
\setlength{\tabcolsep}{1pt}
\caption{\textbf{Axis-flip equivariance ablation.} $k$ denotes the number of anatomical axes enforced ($D$: depth, $H$: height, $W$: width). Training cost per iteration is reported relative to a single-axis baseline (1.18\,s/iter on a Titan V). $^*$: $p < 0.05$ vs.\ $k=3$ under a paired Wilcoxon signed-rank test.}
\label{tab:flip_axes}
\begin{tabular}{clcccr}
\toprule
$k$ & Axes & Dice~$\uparrow$ & HD95 (mm)~$\downarrow$ & ASD (mm)~$\downarrow$ & Cost \\
\midrule
0 & ---             & 85.16$\pm$1.21$^*$ & 13.82$\pm$2.88 & 3.89$\pm$0.74 & 0.81$\times$ \\
\midrule
\multirow{3}{*}{1}
  & $D$            & 85.08$\pm$1.29$^*$ & 12.63$\pm$1.60 & 3.74$\pm$0.70 & 1.02$\times$ \\
  & $H$            & 85.78$\pm$0.88     & 12.30$\pm$1.29 & 3.53$\pm$0.39 & 1.02$\times$ \\
  & $W$            & 85.64$\pm$0.40$^*$ & 12.73$\pm$1.29 & 3.43$\pm$0.32 & 0.96$\times$ \\
\midrule
\multirow{3}{*}{2}
  & $D+H$        & 85.03$\pm$1.06$^*$ & 12.77$\pm$2.14 & 3.91$\pm$0.83 & 1.02$\times$ \\
  & $H+W$        & 86.21$\pm$0.41     & 12.20$\pm$1.23 & 3.34$\pm$0.33 & 1.06$\times$ \\
  & $D+W$        & 85.83$\pm$1.18$^*$ & 12.98$\pm$1.32 & 3.70$\pm$0.58 & 1.06$\times$ \\
\midrule
\rowcolor{gray!25}
3 & $D+H+W$     & \textbf{86.95$\pm$1.21} & \textbf{10.25$\pm$2.61} & \textbf{2.87$\pm$1.13} & 1.26$\times$ \\
\bottomrule
\end{tabular}
\end{table}

\noindent\textbf{Cumulative Component Ablation.} Table~\ref{tab:la_ablation} details step-by-step additions from the supervised baseline under the most constrained setup (4 labeled, 4 unlabeled cases). The full pipeline improves Dice by $+30.46$ points over the baseline ($56.49 \rightarrow 86.95$, $p < 10^{-4}$), while reducing HD95 by $29.24$\,mm and ASD by $1.09$\,mm. The boundary head yields the single largest baseline performance jump ($+9.91$ Dice, $p < 10^{-4}$), followed by mean-teacher consistency ($+7.90$ Dice) and Instance Normalization ($+6.32$ Dice, $p = 0.0014$). Furthermore, seed-to-seed variance contracts monotonically from $\pm 6.71$ to $\pm 1.21$ Dice points as components accumulate, confirming that each module enhances both representation accuracy and optimization stability.

Notably, mean-teacher EMA substantially improves volumetric overlap (+7.90 Dice) while transiently degrading surface metrics (ASD increases from $3.67$ to $5.34$\,mm due to high variance on a single seed). Unfiltered consistency learning across limited unlabeled volumes initially prioritizes coarse regional overlap; adding the uncertainty gate resolves this surface variance, stabilizing ASD back to $3.60$\,mm and reducing HD95 from $21.04$ to $15.42$\,mm.

\begin{table}[!htb]
\centering
\small
\setlength{\tabcolsep}{0.5pt}
\caption{\textbf{Additive component ablation} on the LA dataset (4:4 split; mean~$\pm$~std over 3 seeds). Each row incrementally adds one component to the configuration above.}
\label{tab:la_ablation}
\begin{tabular}{lccc}
\toprule
Component & Dice~$\uparrow$ & HD95 (mm)~$\downarrow$ & ASD (mm)~$\downarrow$ \\
\midrule
Supervised Floor          & 56.49$\pm$4.92 & 39.49$\pm$3.42 & 3.96$\pm$0.50 \\
\quad + Label Smoothing   & 58.72$\pm$6.71 & 36.80$\pm$6.26 & 4.16$\pm$0.44 \\
\quad + Boundary Head     & 68.63$\pm$3.26 & 27.08$\pm$4.88 & 3.67$\pm$0.25 \\
\quad + Mean-Teacher EMA  & 76.53$\pm$2.11 & 21.04$\pm$5.76 & 5.34$\pm$2.60 \\
\quad + Uncertainty Gate & 77.95$\pm$2.04 & 15.42$\pm$0.66 & 3.60$\pm$0.15 \\
\quad + Copy-Paste        & 77.67$\pm$1.12 & 18.28$\pm$0.63 & 4.46$\pm$0.54 \\
\quad + Flip Equivariance & 80.63$\pm$2.11 & 15.85$\pm$1.42 & 3.50$\pm$0.57 \\
\rowcolor{gray!25}
\quad + InstanceNorm      & \textbf{86.95$\pm$1.21} & \textbf{10.25$\pm$2.61} & \textbf{2.87$\pm$1.13} \\
\bottomrule
\end{tabular}
\end{table}

\begin{figure*}
    \centering
    \includegraphics[width=1.0\linewidth]{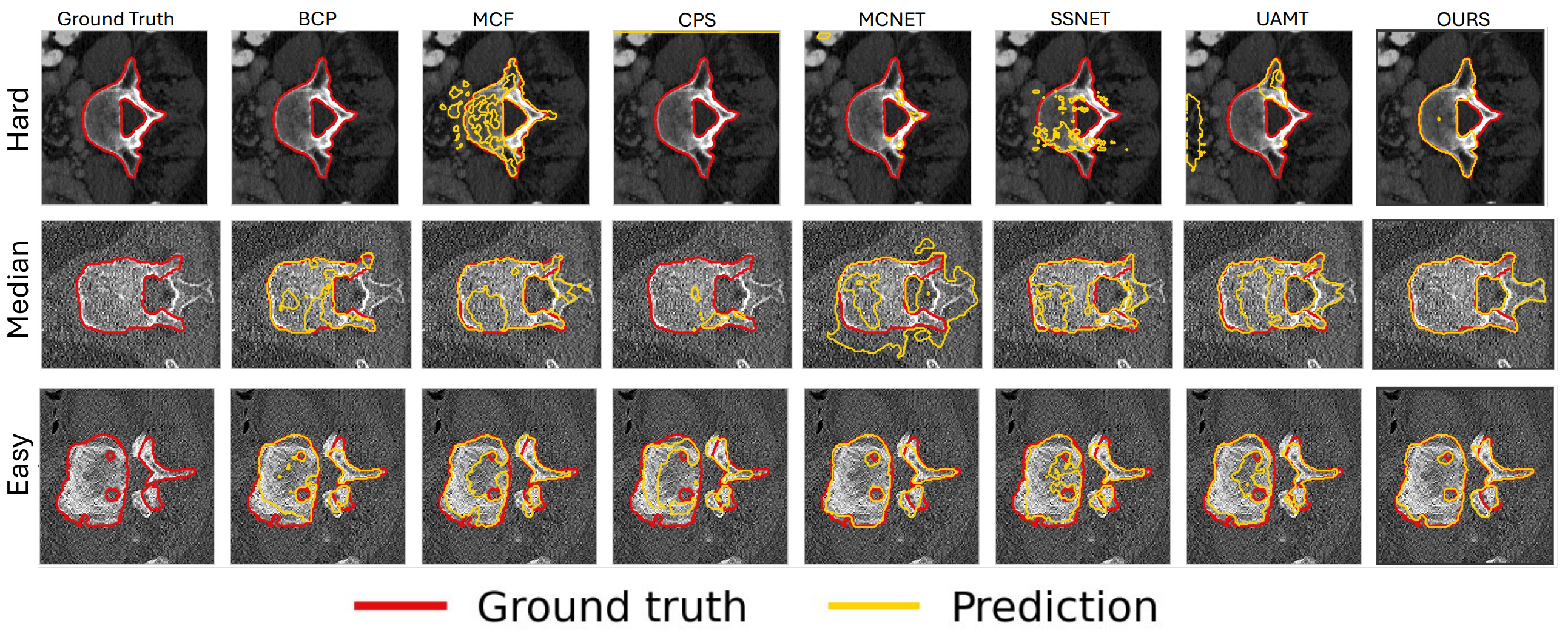}
    \caption{\textbf{Qualitative results on VerSe (5 labeled / 5 unlabeled).} Hardest, median and easiest test cases, selected by mean Dice across all seven methods rather than by our own. Same axial slice per row; red: ground truth, yellow: prediction.}
    \label{fig:verse_qualitative}
\end{figure*}

\begin{figure*}
    \centering
    \includegraphics[width=1.0\linewidth]{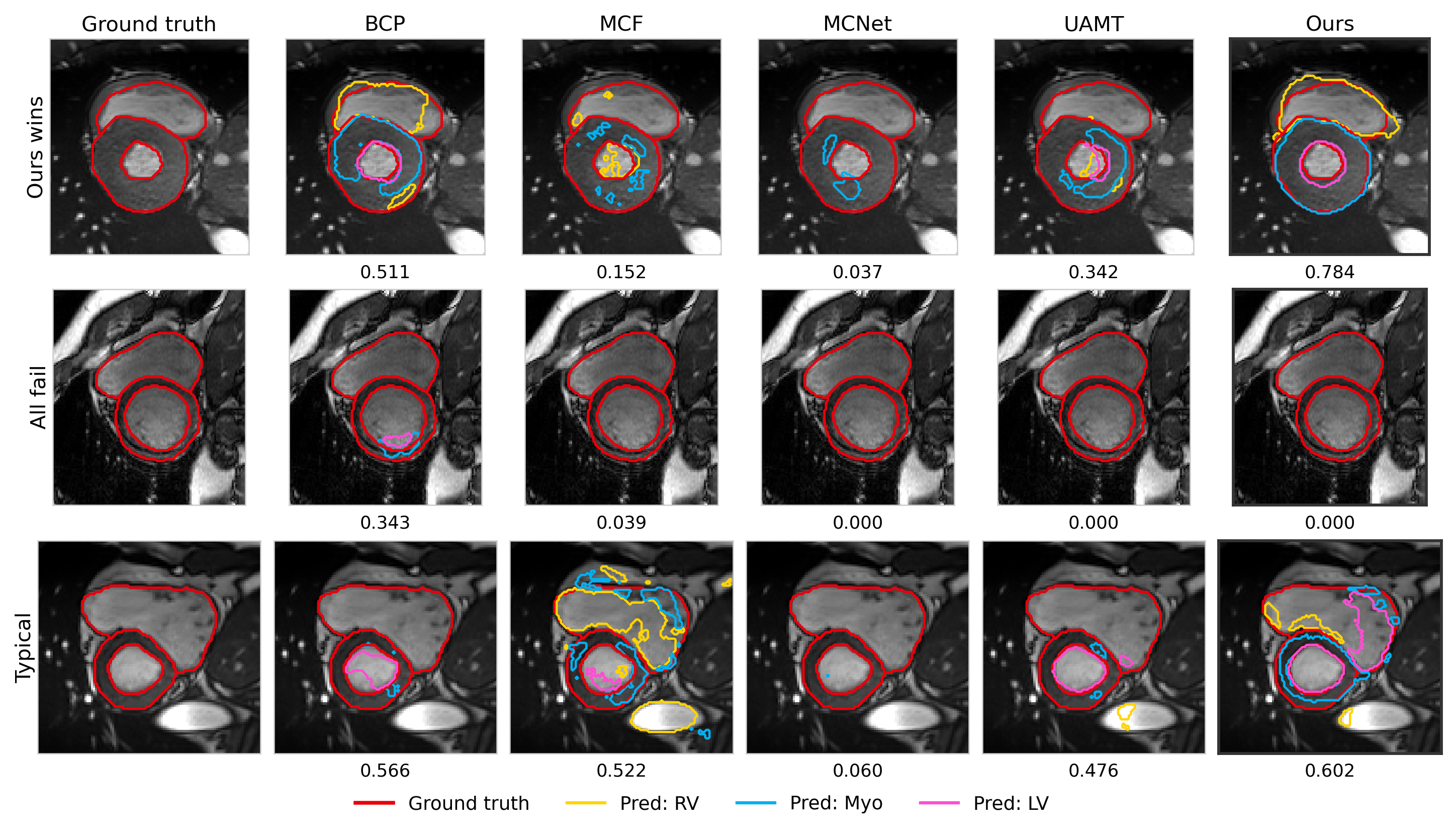}
    \caption{\textbf{Qualitative results on ACDC (7 labeled / 14 unlabeled).} Three representative cases: one where \name clearly outperforms all baselines (selected as the largest margin among cases where every baseline produces a non-empty segmentation), one where every method including ours fails, and one typical case. Unlike Figs.~\ref{fig:la_qualitative} and \ref{fig:verse_qualitative}, the first row is chosen using our own score and is illustrative rather than representative; over the full test set \name outperforms the strongest baseline on $33$ of $40$ cases. Same short-axis slice per row; red: ground truth; predictions coloured by class (yellow: RV, blue: myocardium, pink: LV); values are whole-volume mean Dice}.

    \label{fig:acdc_qualitative}
\end{figure*}

\begin{sidewaystable*}[p]
\centering
\small
\setlength{\tabcolsep}{4pt}
\renewcommand{\arraystretch}{0.90}

\caption{\textbf{LA results at different unlabeled-pool sizes.}
Dice~(\%), HD95~(mm), and ASD~(mm), reported as mean~$\pm$~standard
deviation over three seeds (1337, 42, 123). Column groups indicate
labeled:unlabeled case counts. FS uses no unlabeled data and is therefore
constant across column groups. Best result for each metric and split is
shown in bold.}
\label{tab:la_supp}

\begin{subtable}{\linewidth}
\centering
\caption{Results using four labeled cases.}
\label{tab:la_supp_4lbl}

{%
\begin{tabular}{@{}lccc|ccc|ccc@{}}
\toprule
\multirow{2}{*}{Method}
& \multicolumn{3}{c|}{4:12}
& \multicolumn{3}{c|}{4:28}
& \multicolumn{3}{c}{4:75} \\
\cmidrule(lr){2-4}
\cmidrule(lr){5-7}
\cmidrule(lr){8-10}
& Dice & HD95 & ASD
& Dice & HD95 & ASD
& Dice & HD95 & ASD \\
\midrule
FS
& 58.77$\pm$4.86 & 36.17$\pm$2.99 & 4.96$\pm$1.38
& 58.77$\pm$4.86 & 36.17$\pm$2.99 & 4.96$\pm$1.38
& 58.77$\pm$4.86 & 36.17$\pm$2.99 & 4.96$\pm$1.38 \\
UA-MT\cite{yu2019uncertainty}
& 77.64$\pm$1.24 & 19.18$\pm$1.61 & 4.67$\pm$0.81
& 82.22$\pm$0.24 & 16.68$\pm$2.98 & 5.07$\pm$0.70
& 80.74$\pm$1.53 & 14.97$\pm$2.77 & 4.71$\pm$0.88 \\
DTC\cite{luo2021semi}
& 65.00$\pm$13.73 & 29.15$\pm$12.06 & 4.45$\pm$1.72
& 32.11$\pm$39.89 & 70.39$\pm$51.22 & 12.42$\pm$10.17
& 62.72$\pm$9.88 & 30.58$\pm$9.94 & 5.25$\pm$1.78 \\
SS-Net\cite{wu2022exploring}
& 78.73$\pm$1.60 & 17.51$\pm$1.51 & 3.08$\pm$0.41
& 83.30$\pm$1.53 & 13.38$\pm$1.84 & 2.99$\pm$0.28
& 83.83$\pm$0.74 & 12.69$\pm$0.48 & 2.83$\pm$0.35 \\
BCP\cite{bai2023bidirectional}
& 82.45$\pm$1.07 & 11.92$\pm$0.51 & 2.93$\pm$0.18
& \textbf{87.10$\pm$0.64} & \textbf{7.91$\pm$0.88} & \textbf{2.49$\pm$0.23}
& \textbf{88.43$\pm$1.27} & \textbf{7.35$\pm$1.66} & 2.38$\pm$0.54 \\
CPS\cite{chen2021semi}
& 66.84$\pm$12.33 & 29.94$\pm$8.54 & 6.19$\pm$2.76
& 79.19$\pm$4.15 & 20.93$\pm$6.97 & 6.57$\pm$2.83
& 82.07$\pm$0.23 & 15.74$\pm$2.89 & 4.49$\pm$1.00 \\
MCNet\cite{wu2022mutual}
& 62.52$\pm$26.75 & 26.17$\pm$9.95 & 3.54$\pm$1.53
& 35.87$\pm$31.81 & 28.71$\pm$9.03 & 5.49$\pm$0.96
& 17.94$\pm$29.09 & 18.95$\pm$10.23 & 3.03$\pm$1.28 \\
MCF\cite{wang2023mcf}
& 79.26$\pm$1.37 & 15.50$\pm$3.82 & 4.96$\pm$1.50
& 81.19$\pm$1.30 & 13.99$\pm$2.41 & 4.44$\pm$0.90
& 79.97$\pm$5.19 & 14.00$\pm$4.79 & 4.45$\pm$1.50 \\
CAML\cite{gao2023correlation}
& 80.99$\pm$0.46 & 16.68$\pm$2.31 & \textbf{2.55$\pm$0.04}
& 81.38$\pm$1.48 & 16.70$\pm$1.35 & \bf 2.49$\pm$0.20
& 82.90$\pm$1.52 & 14.32$\pm$2.80 & \textbf{2.34$\pm$0.07} \\
PPC\cite{yuan2024effective}
& 71.39$\pm$3.96 & 25.26$\pm$3.72 & 6.41$\pm$2.51
& 80.35$\pm$2.75 & 18.40$\pm$2.10 & 5.06$\pm$0.63
& 75.85$\pm$5.27 & 21.36$\pm$3.26 & 5.63$\pm$1.34 \\
UnCo\cite{zeng2025uncertainty}
& 70.07$\pm$6.14 & 24.62$\pm$2.26 & 3.84$\pm$0.72
& 78.85$\pm$0.96 & 21.56$\pm$1.50 & 2.74$\pm$0.31
& 78.43$\pm$7.61 & 20.36$\pm$5.79 & 2.61$\pm$0.70 \\
\midrule
\rowcolor{gray!35}
\textbf{Ours}
& \textbf{86.55$\pm$1.16} & \textbf{11.02$\pm$2.05} & 3.21$\pm$0.83
& \bf 86.87$\pm$0.91 & 11.05$\pm$2.27 & 2.94$\pm$0.66
& 86.80$\pm$0.48 & 11.33$\pm$0.61 & 3.17$\pm$0.22 \\
\bottomrule
\end{tabular}%
}
\end{subtable}

\vspace{2mm}

\begin{subtable}{\linewidth}
\centering
\caption{Results using eight labeled cases.}
\label{tab:la_supp_8lbl}

{%
\begin{tabular}{@{}lccc|ccc|ccc@{}}
\toprule
\multirow{2}{*}{Method}
& \multicolumn{3}{c|}{8:8}
& \multicolumn{3}{c|}{8:24}
& \multicolumn{3}{c}{8:71} \\
\cmidrule(lr){2-4}
\cmidrule(lr){5-7}
\cmidrule(lr){8-10}
& Dice & HD95 & ASD
& Dice & HD95 & ASD
& Dice & HD95 & ASD \\
\midrule
FS
& 81.11$\pm$0.67 & 14.97$\pm$1.99 & 2.61$\pm$0.09
& 81.11$\pm$0.67 & 14.97$\pm$1.99 & 2.61$\pm$0.09
& 81.11$\pm$0.67 & 14.97$\pm$1.99 & 2.61$\pm$0.09 \\
UA-MT\cite{yu2019uncertainty}
& 84.47$\pm$0.55 & 12.99$\pm$1.29 & 2.75$\pm$0.58
& 86.69$\pm$0.76 & 11.81$\pm$2.50 & 2.87$\pm$0.59
& 86.68$\pm$1.21 & 10.52$\pm$1.16 & 2.64$\pm$0.17 \\
DTC\cite{luo2021semi}
& 79.95$\pm$3.25 & 18.16$\pm$4.81 & 2.77$\pm$0.51
& 62.77$\pm$25.98 & 36.73$\pm$31.32 & 4.14$\pm$1.90
& 83.78$\pm$2.17 & 12.53$\pm$1.75 & 2.40$\pm$0.31 \\
SS-Net\cite{wu2022exploring}
& 83.85$\pm$0.60 & 13.35$\pm$0.86 & 2.40$\pm$0.01
& 85.01$\pm$0.88 & 12.13$\pm$0.91 & 2.22$\pm$0.08
& 84.41$\pm$1.47 & 12.80$\pm$1.58 & 2.23$\pm$0.17 \\
BCP\cite{bai2023bidirectional}
& 82.39$\pm$2.22 & 11.68$\pm$0.36 & 2.37$\pm$0.16
& 88.08$\pm$0.20 & \bf 7.84$\pm$0.40 & 2.05$\pm$0.01
& \textbf{89.42$\pm$0.11} & \textbf{7.15$\pm$0.36} & \textbf{1.77$\pm$0.07} \\
CPS\cite{chen2021semi}
& 77.10$\pm$7.00 & 17.06$\pm$3.47 & 2.91$\pm$0.48
& 84.30$\pm$4.44 & 12.05$\pm$2.93 & 2.38$\pm$0.23
& 83.48$\pm$3.97 & 14.17$\pm$3.72 & 3.35$\pm$1.37 \\
MCNet\cite{wu2022mutual}
& 85.06$\pm$0.99 & 12.42$\pm$0.89 & \textbf{2.13$\pm$0.09}
& 86.54$\pm$1.09 & 11.48$\pm$1.46 & \textbf{1.94$\pm$0.07}
& 86.96$\pm$0.08 & 10.64$\pm$0.37 & 1.93$\pm$0.04 \\
MCF\cite{wang2023mcf}
& 85.79$\pm$1.28 & \textbf{7.88$\pm$0.44} & 2.37$\pm$0.09
& 86.92$\pm$0.90 & \textbf{7.71$\pm$0.70} & 2.29$\pm$0.21
& 87.23$\pm$0.43 & 7.58$\pm$0.21 & 2.30$\pm$0.09 \\
CAML\cite{gao2023correlation}
& 79.76$\pm$2.38 & 17.17$\pm$2.36 & 2.47$\pm$0.07
& 84.84$\pm$1.69 & 13.47$\pm$2.10 & 2.06$\pm$0.15
& 85.53$\pm$1.01 & 13.23$\pm$1.39 & 2.02$\pm$0.10 \\
PPC\cite{yuan2024effective}
& 83.15$\pm$0.55 & 12.22$\pm$0.76 & 2.56$\pm$0.18
& 86.42$\pm$0.78 & 11.33$\pm$1.88 & 2.92$\pm$0.18
& 86.20$\pm$3.04 & 11.01$\pm$3.59 & 2.89$\pm$0.61 \\
UnCo\cite{zeng2025uncertainty}
& 81.40$\pm$2.09 & 16.97$\pm$2.28 & 2.49$\pm$0.23
& 85.24$\pm$1.26 & 13.59$\pm$2.78 & 2.10$\pm$0.19
& 84.35$\pm$2.14 & 13.93$\pm$2.82 & 1.99$\pm$0.12 \\
\midrule
\rowcolor{gray!35}
\textbf{Ours}
& \textbf{88.85$\pm$0.61} & 9.69$\pm$0.85 & 2.56$\pm$0.29
& \textbf{89.03$\pm$0.64} & 9.01$\pm$2.61 & 2.31$\pm$0.78
& \bf 89.31$\pm$0.09 & \bf 8.25$\pm$0.35 & \bf 2.13$\pm$0.05 \\
\bottomrule
\end{tabular}%
}
\end{subtable}

\end{sidewaystable*}

\end{document}